# A CNN-Based Malaria Diagnosis from Blood Cell Images with SHAP and LIME Explainability

Md. Ismiel Hossen Abir[*1], Awolad Hossain[2]

[1, 2] Department of Computer Science & Engineering,
International Standard University, Dhaka, Bangladesh
{ismielabir286, awoladh04}@gmail.com

**Abstract**

Malaria remains a prevalent health concern in regions with tropical and subtropical climates. The cause of malaria is the Plasmodium parasite, which is transmitted through the bites of infected female Anopheles mosquitoes. Traditional diagnostic methods, such as microscopic blood smear analysis, are low in sensitivity, depend on expert judgment, and require resources that may not be available in remote settings. To overcome these limitations, this study proposes a deep learning-based approach utilizing a custom Convolutional Neural Network (CNN) to automatically classify blood cell images as parasitized or uninfected. The model achieves an accuracy of 96%, with precision and recall scores exceeding 0.95 for both classes. This study also compares the custom CNN with established deep learning architectures, including ResNet50, VGG16, MobileNetV2, and DenseNet121. To enhance model interpretability, Explainable AI techniques such as SHAP, LIME, and Saliency Maps are applied. The proposed system shows how deep learning can provide quick, accurate and understandable malaria diagnosis, especially in areas with limited resources.

# 1. Introduction

Malaria is a severe infectious disease caused by Plasmodium species, which is transmitted to humans via the bites of female Anopheles mosquitoes carrying the parasite [1]. Malaria is commonly found in tropical and subtropical areas around the world [2]. According to the World Malaria Report states that malaria cases rose from 252 million in 2022 to 263 million in 2023 [1]. Therefore, accurate and timely diagnosis is essential for the control and treatment of malaria.

One of the diagnostic methods for malaria is a blood test that detects Plasmodium parasites and identifies the specific species to guide appropriate treatment [2]. But the traditional methods of malaria diagnosis face challenges such as limited sensitivity and dependence on skilled person, which can lead to delayed or inaccurate diagnoses, especially in low-resource settings.

To overcome these issues, we used a deep learning approach to develop a custom Convolutional Neural Network (CNN) to detect malaria from blood cell images. In the dataset contain two classes such as parasitized and uninfected [3]. We also train and compare the performance of several architectures, including ResNet50, VGG16, MobileNetV2, and DenseNet121. In our CNN model, we also include explainable AI techniques such as SHAP (SHapley Additive exPlanations), LIME (Local Interpretable Model-agnostic Explanations), and Saliency Maps to enhance model interpretability [10-12].

The motivation for this work comes from reviewing recent studies, where a lack of explainability analysis. This study highlights the following key contributions:

(i) Developed a custom CNN architecture for malaria detection from blood cell images parasitized and uninfected classes.

(ii) For the model interpretability, implemented explainability SHAP, LIME, and Saliency Maps

This study is organized into several key sections: Section 2 covers the literature review, section 3 methodologies part, section 4 presents the experimental results, section 5 provides the comparative analysis, and section 6 concludes the study, followed by the references.

## 2. Literature Review

Several recent studies have explored deep learning based methods to detect malaria. Maqsood et al. proposed a custom convolutional neural network model for detecting malaria-infected cells from microscopic images, and they achieved 96.82% accuracy score [4]. Bibin et al. proposed a deep belief network (DBN) model for classifying images into parasite and non-parasite classes, and achieved a sensitivity of 97.60%, specificity of 95.92%, and an F-score of 89.66% [5]. Fatima & Farid used a classical image processing approach combining bilateral filtering, adaptive thresholding, and morphological operations, achieving an accuracy of 91.8% on the malaria image dataset [6]. Liang et al. developed a 16-layer CNN model for automatic classification of malaria red blood cells, and they achieved 97.37% accuracy [7]. Quan et al. developed Attentive Dense Circular Net (ADCN) based on CNN, integrating residual, dense connections, and attention mechanisms, and achieved 97.47% accuracy score [8]. Alonso-Ramírez et al. developed CNN models for classifying malaria-infected cells, achieving 97.72% accuracy [9].

## 3. Methodology

This study focuses on a custom Convolutional Neural Network (CNN) for classifying malaria-infected and uninfected blood images. Figure 1 illustrates the overall workflow, which consists of several key phases: data collection, data preprocessing, custom CNN model development, model training and optimization, model evaluation, model interpretation, and a comparative analysis of the proposed CNN. Also, we compared the CNN model performance with four pre-trained models.

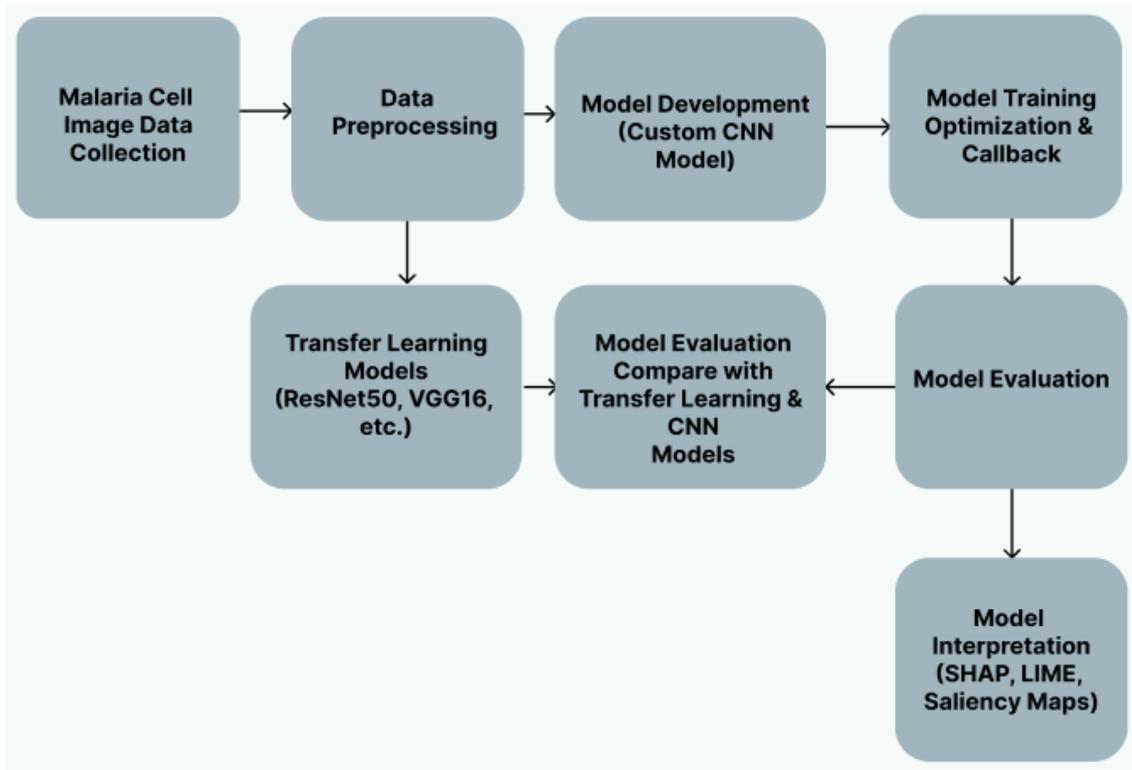

*Figure 1. Workflow Diagram*

### 3.1. Dataset Collection and Preprocessing

The dataset used in this study is sourced from the official repository provided by the National Library of Medicine (NLM), National Institutes of Health (NIH) [3]. Dataset total contains 27558 images, organized into two classes parasitized and uninfected. Each class contains approximately 13,800 images. To enhance model performance, data preprocessing techniques were then applied. In the data preprocessing section steps involve data labeling, class stratification, image resizing, normalization, and data generator preparation. The data was separated into subsets for training, validation, and testing. In the training part used 82% (22596 images) of the data and in the testing, 9% (2481 images), and validation part 9% (2481 images) images used. All the images are resized to $100 \times 100$ pixels, and the RGB channel is used. Image rescaling was used by dividing by 255. The experiments were conducted in the Kaggle coding environment, utilizing a NVIDIA Tesla P100 GPU for accelerated training. A batch size of 32 was used for all data generators to balance memory efficiency and model convergence. During preprocessing, class labels were encoded as parasitized=0 and uninfected=1. Figure 2, presents the sample image from the dataset. The left cell is uninfected, showing a normal red blood cell structure. On the other hand, the right cell is parasitized, identifiable by the presence of a distinct purple-stained Plasmodium parasite.

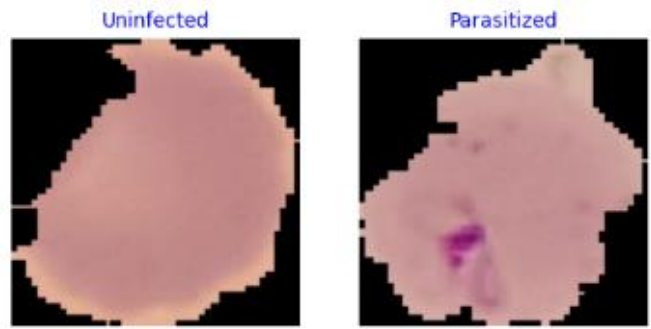

*Figure 2.* Dataset Visualization

### *3.2. Custom CNN Model Architecture*

For binary classification of parasitized and uninfected classes, we used custom CNN model. Figure 3 represent the custom CNN model architecture. In this architecture, we used four main convolutional block. First three convolutional blocks we used two convolutional layers and for the fourth block used one convolutional layer. Each of the convolutional layers contains the ReLU as activation function and 3X3 kernel size. The number of filters increases with each block 32, 64, 128, and 256 to allow the network to learn both low-level and high-level features. Each of the convolutional blocks we used Batch Normalization layer to stabilize and speed up the training process. In the first three convolutional blocks, we used Max Pooling layer with (2, 2) pool size to reduce the size of the feature maps and help the model focus on the most important patterns. In the final convolutional block, a Global Average Pooling layer is used. Then, a Dropout layer with a 0.25 rate was included to prevent overfitting.. After that, we used a flatten and three fully connected layer. Two fully connected layers with 128 and 64 neurons used. Both layers use ReLU activation and L2 regularization (0.01) to improve generalization. After that final fully connected layer for binary classification with sigmoid activation function. The model contains a total of 625,378 parameters. We also used learning rate scheduling techniques Reduce LR On Plateau. To monitor the training, we used a custom callback function that stops training early if the model reaches 99% accuracy on either training or validation data. This helps save time and avoids overfitting. The model was compiled using the Adamax optimizer with an initial learning rate of 0.001.

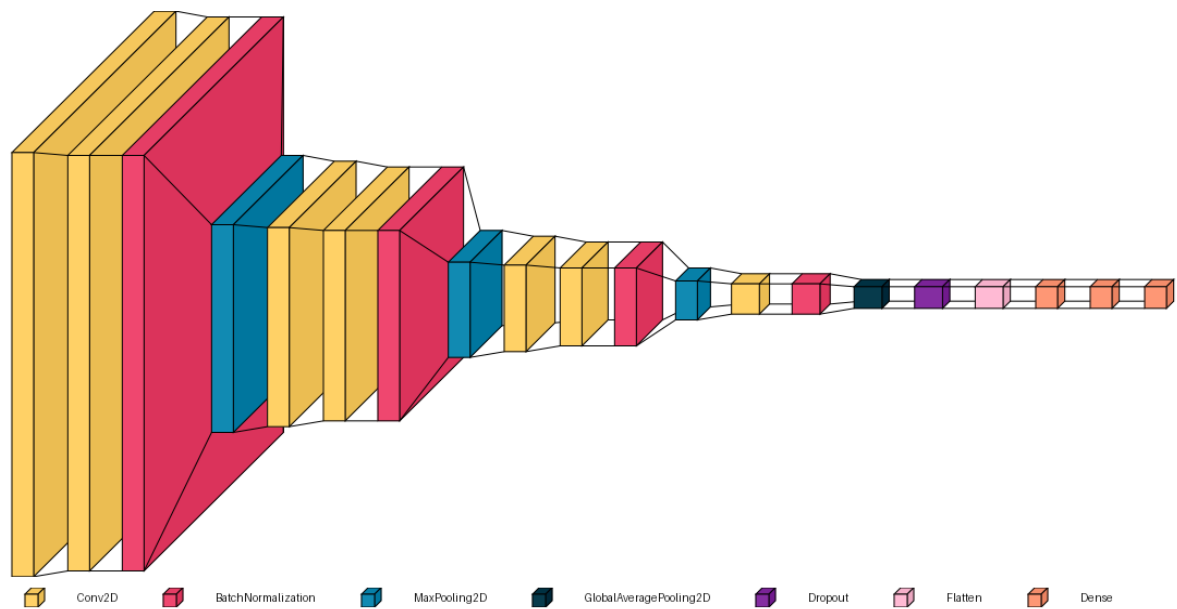

*Figure 3.* Custom CNN Model Architecture

### *3.3. Model Evaluation and Explainable AI*

After training the custom CNN model, several evaluation metrics and model interpretation techniques were employed, including the classification report, confusion matrix, SHAP (SHapley Additive exPlanations), LIME (Local Interpretable Model-Agnostic Explanations), and Saliency Maps [10-12]. Firstly, we used classification report which includes evaluation metrics such as accuracy, precision, recall, and F1-score. The classification report provides detailed about the model's ability to correctly classify parasitized and uninfected class. After that, we used confusion matrix that represent the models identification true positives, true negatives, false positives, and false negatives. That helps to identify the model prediction and errors. After that, we used explainable AI techniques SHAP, LIME, and Saliency Maps. SHAP and LIME provided pixel level and region specific insights into model behavior, demonstrating the model's transparency and confirming that each class was correctly identified [10,11]. Saliency Maps were used as a gradient based approach to highlight the most influential image regions [12].

### *3.4. Transfer Learning Model*

To compare with our custom CNN model, we used four deep learning models through transfer learning: ResNet50, VGG16, MobileNetV2, and DenseNet121 [13-16]. Each of these models was originally trained on the large ImageNet dataset. In the transfer learning models, we reused the features and applied them to our malaria image classfication task. For each model, we removed the top classification layers and kept only the convolutional base by using include_top=False. Then, set the input image size to 100×100×3 to match our dataset. We also added a few new layers to each base model. First, we used a Global Average Pooling layer to reduce the number of features. After that, we added a Dense layer with 128 neurons and the ReLU activation function. To reduce overfitting, include a Dropout layer with a dropout rate of 0.5. Finally, added a Dense layer with 2 neurons and a softmax activation to make predictions for the two classes: parasitized and uninfected. We used the Adam optimizer to train the models and set the loss function to categorical cross-entropy [17]. This helped use transfer learning models for malaria detection and compare them with custom CNN.

### 4. Experimental Result

In this section, we divided it into model evaluation metrics results and explainable AI.

## 4.1. Model Evaluation Metrics Analysis

We used classification report and confusion matrix to observe custom CNN model evolution matrix. Table 1, represent classification matrix. The model achieved overall 96% accuracy score. Precision, recall, and f1-score range is 95% - 97% both of the classes. The parasitized class precision and f1-score us 96% and recall is 95%. On the other hand, in the uninfected class precision 95%, recall 97% and f1-score 96%. Overall, the scores indicated that model's strong and balanced performance across both classes. Figure 4 represents confusion matrix. The model correctly classified 1174 parasitized and 1197 uninfected cell images. Though the model misclassified 67 parasitized and 43 uninfected classes but the number of misclassifications is much lower than correctly classified.

*Table 1: Classification Report*

|  | Precision | Recall | F1-score |
|---|---|---|---|
| Parasitized | 0.96 | 0.95 | 0.96 |
| Uninfected | 0.95 | 0.97 | 0.96 |
| Accuracy |  |  | 0.96 |

## 4.2. Model Explainability Analysis

In this section, SHAP, LIME, and saliency maps are used to interpret the custom. Figure 5, represent LIME of both classes. Figure 5, subfigure (a) represent parasitized class original image and subfigure (b) represent LIME of the corresponding image. In the subfigure (b) magenta colored regions indicate features that contributed most significantly to the model's classification and based on this it identify it as parasitized class. On the other hand, subfigure (c) represent distribution and assess load transfer through the dowel. Original image of uninfected class and subfigure (d) represent its LIME image where it shows no significant infection and identify as uninfected class. Both subfigure (b) and (d) yellow highlighted region mainly for original images boundaries.

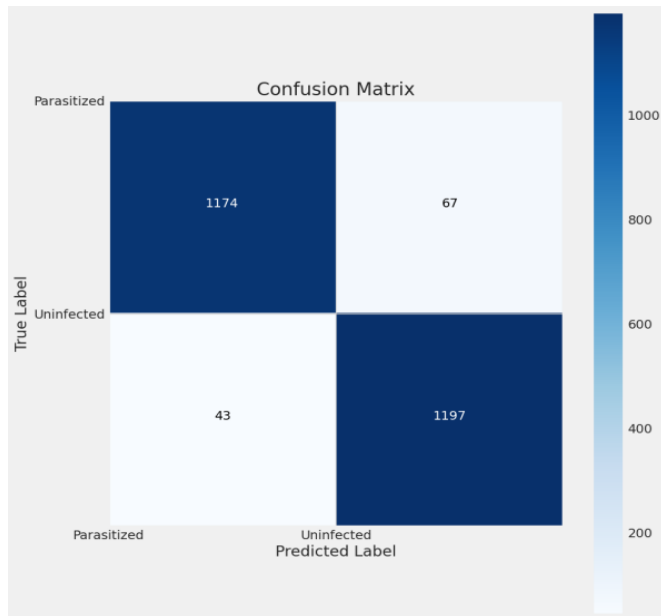

Figure 4. Confusion Matrix

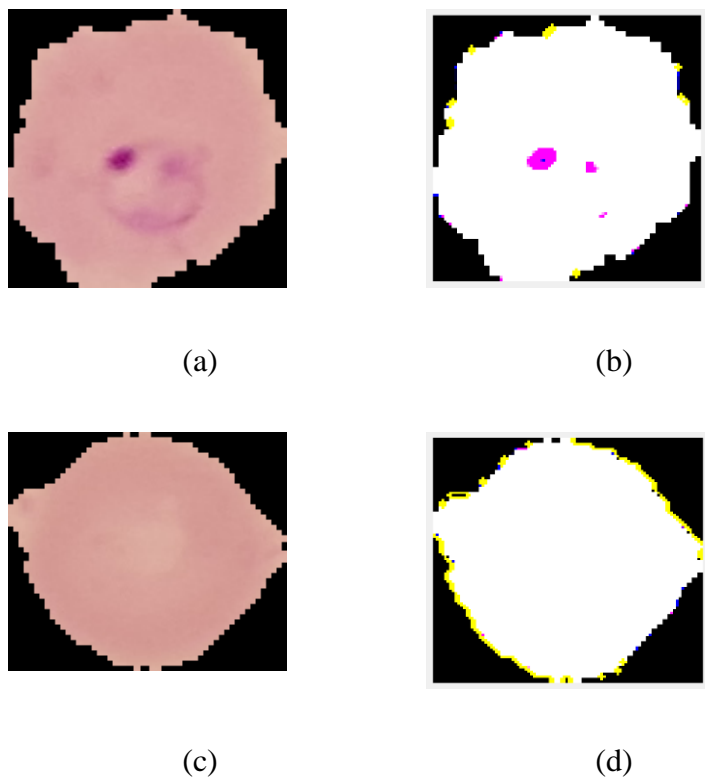

Figure 5. (a) Original parasitized cell image. (b) Corresponding LIME explanation
(c) Original uninfected cell image (d) Corresponding LIME explanation

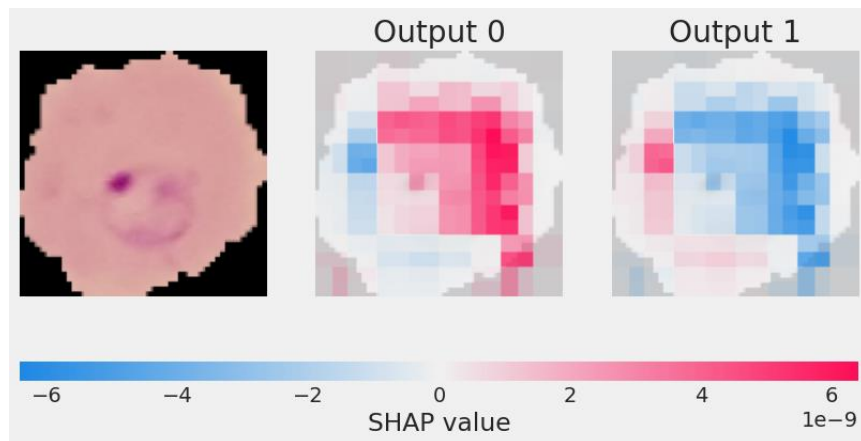

Figure 6. Left: Original parasitized cell image; Middle: SHAP for Output 0 (Parasitized class); Right: SHAP for Output 1 (Uninfected class)

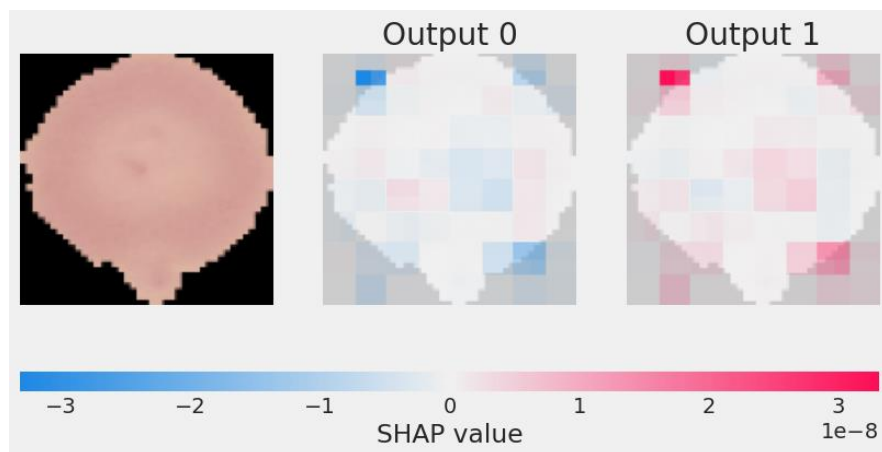

Figure 7. Left: Original uninfected cell image; Middle: SHAP for Output 0 (Parasitized class); Right: SHAP for Output 1 (Uninfected class)

Both figure 6 and 7 represent SHAP of both parasitized and uninfected class. Both figure 6 and 7, left image represented original images, middle image for parasitized class and right image for uninfected class. In the figure 6 middle image shows red color that indicate it positively influence the classification as parasitized and right image shows mostly blue color that indicate negatively identify. So, figure 6 SHAP identify it as parasitized class and the original image also parasitized class. Similarly, figure 7, red region indicate positively of the class and blue indicate negatively of the class. That means figure 7 indicates uninfected class and original image also uninfected class. So, SHAP transparency represents the classification of the model. Based on Figure 8, the left side image represents the original parasitized cell image and the right image its saliency map. The right image highlights the most important region used by the model to detect infection. The bright spot in the center indicates where the model focused to classify the cell as parasitized.

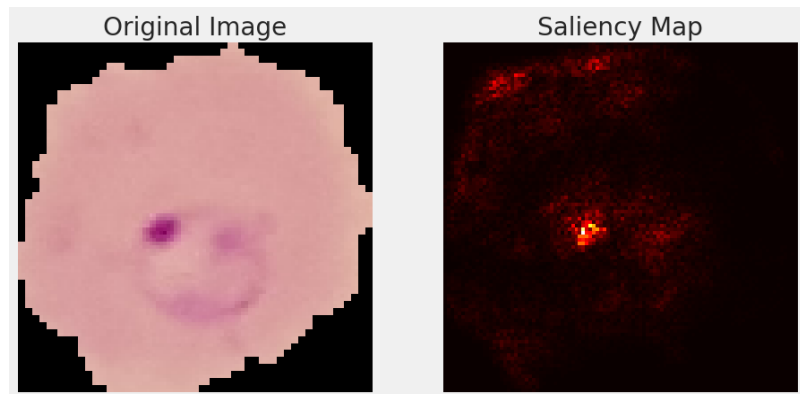

Figure 8. Saliency maps for Parasitized class

## 5. Comparative Analysis

In this section, we compare the performance of our custom CNN model with 4 transfer learning models and recent malaria detection studies from the literature. Table 2, compare with 4 transfer learning model. The models, VGG16 achieved the highest accuracy of 95.28%, ResNet50 and DenseNet121, with scores of 92.30% and 91.77%, respectively. MobileNetV2, which is optimized for mobile devices, achieved a lower accuracy of 80.85%. But the custom CNN model developed in this research outperformed all standard models, achieving an accuracy of 96%. Also maintaining lower computational complexity, total params 625,378 (2.39 MB). Table 3, represents a comparison between the proposed model and existing studies in malaria detection. Though some previous works have achieved slightly higher accuracy but they do not use Explainable AI. The proposed model balances high accuracy 96%, interpretability through SHAP, LIME, and Saliency maps.

*Table 2. Performance Comparison with Transfer Learning Model*

| Model Name | Accuracy Score (%) |
|---|---|
| ResNet50 | 92.30 |
| VGG16 | 95.28 |
| MobileNetV2 | 80.85 |
| DenseNet121 | 91.77 |
| Custom CNN Model developed in this research | 96 |

*Table 3. Comparative Analysis*

| References | Methodology | Accuracy Score (%) | Explainable AI |
|---|---|---|---|
| [5] | CNN | 96.82 | No |
| [7] | Classical image processing (bilateral filtering + adaptive thresholding + morphological operations) | 91 | No |
| [8] | CNN | 97.37 | No |
| [9] | Attentive Dense Circular Net (ADCN) with attention and residual | 97.47 | No |
| [10] | CNN | 97.72 | No |
| Model developed in this research | Custom CNN | 96 | Yes (SHAP, LIME, Saliency maps) |

## 6. Conclusions

This research presents a deep learning-based solution for malaria diagnosis using blood cell images. A custom Convolutional Neural Network (CNN) was developed and achieved a high accuracy of 96%. Our CNN model represents better performance than several transfer learning models such as ResNet50, VGG16, DenseNet121, and MobileNetV2. Also, the model

Page 12 of 14

incorporates explainable AI techniques, SHAP, LIME, and Saliency Maps to provide transparency and visual insights into the decision-making process. This level of interpretability enhances user trust and ensures transparency, particularly essential in medical applications. We also compared to previous studies and many of which report slightly higher accuracy scores but our proposed model uniquely combines strong performance and model explainability. Model transparency and practical implementation in medical images are essential. Overall, this research not only advances diagnostic accuracy but also ensures clinical trust through explainable AI.

## References

1. World Health Organization. Malaria. World Health Organization https://www.who.int/news-room/fact-sheets/detail/malaria (2024).

2. Cleveland Clinic. Malaria. Cleveland Clinic Health Library https://my.clevelandclinic.org/health/diseases/15014-malaria (2025).

3. National Library of Medicine. Malaria datasets. National Institutes of Health https://ceb.nlm.nih.gov/repositories/malaria-datasets/ (2018).

4. Maqsood, A., Farid, M. S., Khan, M. H. & Grzegorzek, M. Deep malaria parasite detection in thin blood smear microscopic images. Appl. Sci. 11, 2284 (2021).

5. Bibin, D., Nair, M. S. & Punitha, P. Malaria parasite detection from peripheral blood smear images using deep belief networks. IEEE Access 5, 9099–9108 (2017).

6. Fatima, T. & Farid, M. S. Automatic detection of Plasmodium parasites from microscopic blood images. J. Parasit. Dis. 44, 69–78 (2020).

7. Liang, Z. et al. CNN-based image analysis for malaria diagnosis. In Proc. IEEE Int. Conf. Bioinformatics and Biomedicine 493–496 (IEEE, 2016).

8. Quan, Q., Wang, J. & Liu, L. An effective convolutional neural network for classifying red blood cells in malaria diseases. Interdiscip. Sci. Comput. Life Sci. 12, 217–225 (2020).

9. Alonso-Ramírez, A.-A. et al. Malaria cell image classification using compact deep learning architectures on Jetson TX2. Technologies 12, 247 (2024).

10. Lundberg, S. M. & Lee, S.-I. A unified approach to interpreting model predictions. In Advances in Neural Information Processing Systems 30, 4765–4774 (2017).

11. Ribeiro, M. T., Singh, S. & Guestrin, C. "Why should I trust you?": Explaining the predictions of any classifier. In Proc. 22nd ACM SIGKDD Int. Conf. Knowledge Discovery and Data Mining 1135–1144 (ACM, 2016).

12. Simonyan, K., Vedaldi, A. & Zisserman, A. Deep inside convolutional networks: Visualising image classification models and saliency maps. In ICLR Workshop (2014).

13. He, K., Zhang, X., Ren, S. & Sun, J. Deep residual learning for image recognition. In Proc. IEEE Conf. Computer Vision and Pattern Recognition 770–778 (IEEE, 2016).

14. Simonyan, K. & Zisserman, A. Very deep convolutional networks for large-scale image recognition. In Int. Conf. Learning Representations (2015).

15. Sandler, M. et al. MobileNetV2: Inverted residuals and linear bottlenecks. In Proc. IEEE Conf. Computer Vision and Pattern Recognition 4510–4520 (IEEE, 2018).

16. Huang, G., Liu, Z., Van Der Maaten, L. & Weinberger, K. Q. Densely connected convolutional networks. In Proc. IEEE Conf. Computer Vision and Pattern Recognition 4700–4708 (IEEE, 2017).

17. Kingma, D. P. & Ba, J. Adam: A method for stochastic optimization. In Int. Conf. Learning Representations (2015).